%% file: acl2023.tex
\pdfoutput=1
\documentclass[11pt]{article}

\usepackage[]{ACL2023}
\usepackage{times}
\usepackage{latexsym}

\usepackage[T1]{fontenc}
\usepackage[utf8]{inputenc}

\usepackage{microtype}
\usepackage{inconsolata}

\usepackage{booktabs}
\usepackage{multirow}
\usepackage{tabularx}
\usepackage{tabu}
\usepackage[english]{babel}
\usepackage{microtype}
\usepackage{lipsum}
\usepackage{hyperref}
\usepackage{tabularray}
\usepackage{graphics}

\usepackage{enumitem}
\setlist{nolistsep}

\usepackage{xspace}
\newcommand{\modelname}{\textsc{CamemBERTa}\xspace}

\title{Data-Efficient French Language Modeling with \textsc{CamemBERTa}}

\author{Wissam Antoun \quad Benoît Sagot \quad Djamé Seddah \\
     Inria, Paris\\
     \{firstname,lastname\}@inria.fr}

\begin{document}
\maketitle
\input{sections/abstract.tex}

\input{sections/intro.tex}

\input{sections/lit_review}
\input{sections/methodology}
\input{sections/experiments}
\input{sections/discussion}
\input{sections/conclusion}
\input{sections/extra}

% Entries for the entire Anthology, followed by custom entries
\bibliography{anthology,custom}
\bibliographystyle{acl_natbib}

\clearpage
\onecolumn
\appendix

\input{sections/appendix}

\end{document}

%% file: sections/abstract.tex
\begin{abstract}
    Recent advances in NLP have significantly improved the performance of language models on a variety of tasks.
    While these advances are largely driven by the availability of large amounts of data and computational power, they also benefit from the development of better training methods and architectures.
    In this paper, we introduce \modelname, a French DeBERTa model that builds upon the DeBERTaV3 architecture and training objective.
    We evaluate our model's performance on a variety of French downstream tasks and datasets, including question answering, part-of-speech tagging, dependency parsing, named entity recognition, and the FLUE benchmark, and compare against CamemBERT, the state-of-the-art monolingual model for French.
    Our results show that, given the same amount of training tokens, our model outperforms BERT-based models trained with MLM on most tasks.
    Furthermore, our new model reaches similar or superior performance on downstream tasks compared to CamemBERT, despite being trained on only 30\% of its total number of input tokens.
    In addition to our experimental results, we also publicly release the weights and code implementation of \modelname, making it the first publicly available DeBERTaV3 model outside of the original paper and the first openly available implementation of a DeBERTaV3 training objective.\footnote{\href{https://gitlab.inria.fr/almanach/CamemBERTa}{https://gitlab.inria.fr/almanach/CamemBERTa}}
\end{abstract}

%% file: sections/intro.tex
\section{Introduction}
Advances in natural language processing (NLP) have been driven mainly by scaling up the size of pre-trained language models, along with the amount of data and compute required for training~\cite{raffel2019exploring,radford2019language,rae2021scaling,fedus2021switch,hoffmann2022training}.
However, these are not the only factors to determine a model's downstream performance, as the model's architecture and training objective are also important.
\citet{he2021deberta} showed that we can improve a model's performance by using disentangled attention, which uses two vectors to represent a token, one for position and one for content.
\citet{he2021debertav3} later showed that performance could be further improved by using ELECTRA's~\cite{clark2020electra} self-supervised and sample-efficient replaced token detection objective.
Another crucial aspect lies in the ability to train models faster, which allows for quick iteration and thus accelerates the research process and allows for more efficient exploration of new ideas \cite{izsak-etal-2021-train,extreme-bert,geiping2022cramming}.

This research aims to develop data-efficient and optimized training techniques that can improve performance in downstream tasks, while reducing the required training corpus size and compute.
To achieve this goal, we propose a new data-efficient French language model based on DeBERTaV3~\cite{he2021debertav3}.
Our proposed model aims to optimize the training process by using a sample-efficient training objective, a state-of-the-art model architecture, and an efficient implementation.
We evaluate downstream performance with a variety of NLP tasks, including dependency parsing, part-of-speech tagging, named entity recognition, text classification, and question answering.
We compare our model to a BERT model trained with the masked language modeling (MLM) objective using the same tokenizer and training corpus, and to the state-of-the-art French language model, CamemBERT~\cite{martin-etal-2020-camembert}, which required three times as many training iterations.
Our results show that our proposed model reaches or establishes a new state-of-the-art using one third of the computational budget of its main predecessors.

Our contributions can be summarized as follows:

\begin{itemize}
\item We propose a new data-efficient French language model, which we train based on our DeBERTaV3 re-implementation with our optimized training recipe.
\item We empirically show that under the same conditions, our model outperforms Transformer models trained with MLM on most tasks, and that it reaches or establishes a new state-of-the-art even when compared with models trained for three times as long.
\item Our release is the only publicly available implementation of DeBERTaV3's training objective, and the first for a monolingual \mbox{DeBERTaV3} model other than the original paper.
\end{itemize}

Our code and models are available under an open-source license\footnote{\href{https://gitlab.inria.fr/almanach/CamemBERTa}{https://gitlab.inria.fr/almanach/CamemBERTa}}, making it easy for researchers to reproduce our results and build upon our work.

%% file: sections/lit_review.tex
\section{Related Works}
\label{sec:related-works}
\paragraph{Transformers.}  This architecture has been widely adopted in NLP tasks such as language modeling, mainly due to the use of the self-attention mechanisms~\cite{vaswani2017attention}, which allow the model to weigh the importance of different parts of the input when making predictions.
A downside of the Transformer block is that it is permutation-invariant, which inhibits the model from encoding word order information.
Originally, the authors proposed to add either a fixed sinusoidal pattern or a learned positional embedding as positional bias the input token embedding.
Later studies have shown that using relative positional embeddings is more effective~\cite{shaw-etal-2018-self,dai-etal-2019-transformer,qu-etal-2021-explore}.
Recently, \citet{he2021deberta} proposed a new disentangled attention mechanism, which considers both the relative position and the content of the input tokens as separate vectors.

\paragraph{Pre-trained French Language Models.}
Current language models available for French are either trained using Masked Language Modeling (MLM) or Causal Language Modeling (CLM).
\mbox{CamemBERT} \cite{martin-etal-2020-camembert} and \mbox{FlauBERT} \cite{le-etal-2020-flaubert-unsupervised} are two of the most popular contemporary French models, both trained with masked language modeling.
Other models include FrALBERT \cite{cattan:hal-03336060}, a French version of ALBERT~\cite{Lan2020ALBERT}, LePetit~\cite{micheli-etal-2020-importance} which is a small version of CamemBERT, and D’AlemBERT~\cite{gabay-etal-2022-freem}, a RoBERTa~\cite{liu2020roberta} based language model targeted towards Early Modern French. BARThez~\cite{kamal-eddine-etal-2021-barthez} is a sequence-to-sequence model trained with BART's objective ~\cite{lewis-etal-2020-bart}, and PAGnol~\cite{launay-etal-2022-pagnol} and Cedille~\cite{muller2022cedille} are models trained with the CLM objective.

To the best of our knowledge, there is no prior effort in developing language models with this improved disentangled attention mechanism and objectives other than MLM/CLM beyond English.

%% file: sections/methodology.tex
\section{\modelname: Methodology}
The following section details our proposed architecture and pre-training objective, along with descriptions for the downstream tasks.

\paragraph{Architecture}
\modelname is based on the DeBERTaV3~\cite{he2021deberta} architecture which uses two vectors to encode the word  and its position, with the premise being that the relative position of a word pair should also directly affect the computed attention weights.
The V3 version optimizes the initial DeBERTa architecture by sharing the relative position embedding projection layers across all the encoder layers, and by adding a convolution layer aside the first encoder layer.\footnote{See Section 5.3 of the DeBERTa paper~\cite{he2021deberta}}
We use a base model configuration with 12 layers and 12 attention heads, 768 hidden dimensions with 32k for vocabulary size.

\paragraph{Training Objective}
We follow the DeBERTaV3~\cite{he2021debertav3} pretraining strategy by using the replaced token detection (RTD) pre-training loss first introduced in ELECTRA~\cite{clark2020electra}, with a generator and discriminator based on the DeBERTa architecture.
During pre-training we project the generator embeddings to 256 dimensions and keep the generator model at 12 layers.

During pre-training the generator model is trained using the MLM objective where we dynamically mask 15\% of the input tokens.
We then sample from the generator the masked tokens, and feed the output along with the unmasked tokens to the discriminator which is tasked to identify tokens that were replaced by the generator.
The RTD objective increases sample efficiency since the model is predicting over all input tokens instead of the 15\% masked tokens.

In DeBERTaV3, the authors hypothesized and showed that sharing token embeddings between the generator and the discriminator results in a tug-of-war situation, where the MLM and RTD tasks pull the embedding vectors into opposing directions.
To alleviate this problem, the authors implemented Gradient-Disentangled Embedding Sharing (GDES), a method that re-parameterize the discriminator's token embeddings as $E_D = sg(E_G) + E_\Delta$, where $sg$ stops the gradient flow from the RTD loss to the generator token embeddings $E_G$, and hence the loss gradient only updates a Difference Embedding matrix $E_\Delta$ that is added to $E_G$ to form the discriminator token embeddings $E_D$.
After pre-training, $E_\Delta$ and $E_G$ are summed to get the final $E_D$ and $E_\Delta$ is then discarded.

\paragraph{Pre-Training}
We pre-train on the French subset of \mbox{CCNet}\footnote{See Appendix~\ref{app:ccnet} for more information on dataset choice.}~\cite{wenzek-etal-2020-ccnet}, the same corpus used to pre-train CamemBERT$_{CCNet}$~\cite{martin-etal-2020-camembert}.\footnote{We go over the pertaining dataset choice in the experiments section.}
Moreover we reuse CamemBERT$_{CCNet}$'s  tokenizer \cite{kudo2018sentencepiece}.
By reusing the pre-training corpus and tokenizer,  we isolate the performance differences to the model architecture and training objective variables.

\paragraph{Optimization}
To speed up the pre-training experiments, we split the pre-training into two phases; in phase 1, the model is trained with a maximum sequence length of 128 tokens for 10,000 steps with 2,000 warm-up steps and a very large batch size of 67,584.
In phase 2, maximum sequence length is increased to the full model capacity of 512 tokens for 3,300 steps with 200 warm-up steps and a batch size of 27,648.
Because we use very large batch sizes, we optimize the model using the LAMB optimizer~\cite{You2020Large} with a learning rate of $6e^{-3}$, $\beta_1 = 0.878$, and $\beta_2 = 0.974$.

%% file: sections/experiments.tex
\section{Experiments and Results}

\paragraph{Pre-Training Setup}
We re-implement the \mbox{DeBERTaV3} RTD pre-training objective with GDES, since no public implementation was available at the time of writing.
Our training implementation is based on Nvidia's ELECTRA and BERT TensorFlow2 implementations.\footnote{\href{https://github.com/NVIDIA/DeepLearningExamples/tree/master/TensorFlow2/LanguageModeling/}{https://github.com/NVIDIA/DeepLearningExamples/}}
We train our models for 8 days on 6 Nvidia A40 with Horovod~\cite{sergeev2018horovod}, and make use of XLA compilation, mixed-precision and gradient accumulation to speed-up training and to fit large batch sizes with our limited compute.

During pre-training, our model would have seen 133B tokens compared to 419B tokens for CamemBERT$_{CCNet}$ which was trained for 100K steps. This represents roughly 30\% of CamemBERT's full training.
Hence for a fair comparison, we train a RoBERTa model, which we dub CamemBERT$_{30\%}$, using our same exact pre-training setup but with the MLM objective.

\paragraph{Downstream Evaluation}
We compare our models, CamemBERT$_{CCNet}$, and CamemBERT$_{30\%}$, on a diverse set of French downstream tasks and datasets, namely:
Question Answering (QA) on FQuAD 1.0~\cite{2020arXiv200206071}, Part-Of-Speech (POS) tagging and Dependency Parsing on GSD~\cite{mcdonald-etal-2013-universal}, Rhapsodie~\cite{lacheret:halshs-01061368}, Sequoia~\cite{candito-seddah-2012-le,CANDITO14.494} in their UD v2.2 versions and the French Social Media Bank\footnote{We follow \citet{riabi-etal-2021-character} and use their shuffled version of the treebank, which they split into around 2000 sentences for training, and 1000 for each the dev and test sets}~\cite{seddah-etal-2012-french}, Named Entity Recognition (NER) on the 2008 version of FTB~\cite{abeille-etal-2000-building,candito-crabbe-2009-improving} with NER annotation by \citet{sagot-etal-2012-annotation}, and the FLUE benchmark~\cite{le-etal-2020-flaubert-unsupervised}.

We use the dataset splits as provided by their respective authors, and we finetune using well-tested scripts from the Hugging Face \mbox{\em Transformers} library and the HOPS parser~\cite{grobol:hal-03223424}.
We only perform hyper-parameter tuning for the NER and QA tasks. See Appendix~\ref{sec:HP-appendix} for task-specific details.
\textbf{Bold} text shows the best statistically significant score over 5 seeds.

\begin{table*}[ht!]

    \centering
    {\footnotesize
\resizebox{\textwidth}{!}{%
        \begin{tabu}{ l  c  c @{\hspace{0.35cm}}  @{\hspace{0.35cm}} c  c @{\hspace{0.35cm}}  @{\hspace{0.35cm}} c  c  @{\hspace{0.35cm}}  @{\hspace{0.35cm}} c  c |c}
            \toprule
                                                  & \multicolumn{2}{c @{\hspace{0.5cm}}}{\textsc{GSD}} & \multicolumn{2}{c @{\hspace{0.7cm}}}{\textsc{Rhapsodie}} & \multicolumn{2}{c @{\hspace{0.7cm}}}{\textsc{Sequoia}} & \multicolumn{2}{c @{\hspace{0.35cm}}}{\textsc{FSMB}} & {\sc NER}\\
            \cmidrule(l{2pt}r{0.4cm}){2-3}\cmidrule(l{-0.2cm}r{0.4cm}){4-5}\cmidrule(l{-0.2cm}r{0.4cm}){6-7}\cmidrule(l{-0.2cm}r{2pt}){8-9}\cmidrule(l{0cm}r{1pt}){10-10}
            \multirow{-2}{*}[1pt]{\textsc{Model}} & \textsc{UPOS}                                      & \textsc{LAS}                                             & \textsc{UPOS}                                          & \textsc{LAS}                                         & \textsc{UPOS}  & \textsc{LAS}   & \textsc{UPOS}  & \textsc{LAS} & {\sc F1}  \\
            \midrule
            CamemBERT$_{30\%}$                   & 98.55{\scriptsize$\pm$0.05}  & \textbf{94.26}{\scriptsize$\pm$0.03}     & \textbf{97.61}{\scriptsize$\pm$0.12}   & 83.19{\scriptsize$\pm$0.62}     & \textbf{99.32}{\scriptsize$\pm$0.08}          & 94.09{\scriptsize$\pm$0.06}          & 94.63{\scriptsize$\pm$0.11}          & 80.13{\scriptsize$\pm$0.41}    &   \textbf{91.04}{\scriptsize$\pm$0.76}    \\
            
             CamemBERT$_{CCNet}$                   & \textbf{98.57}{\scriptsize$\pm$0.07} & \textbf{94.35}{\scriptsize$\pm$0.15}    & \textbf{97.62}{\scriptsize$\pm$0.08}    & \textbf{84.29}{\scriptsize$\pm$0.56}        & \textbf{99.35}{\scriptsize$\pm$0.09}          & \textbf{94.78}{\scriptsize$\pm$0.12}          & \textbf{94.80}{\scriptsize$\pm$0.16} & \textbf{81.34}{\scriptsize$\pm$0.63}  & 89.97{\scriptsize$\pm$0.50} \\ %
            \midrule
            \modelname                            & \textbf{98.55}{\scriptsize$\pm$0.05}   & \textbf{94.38}{\scriptsize$\pm$0.15}       & \textbf{97.52}{\scriptsize$\pm$0.14} & \textbf{84.23}{\scriptsize$\pm$0.08}                                                & \textbf{99.44}{\scriptsize$\pm$0.07} & \textbf{94.85}{\scriptsize$\pm$0.14} & \textbf{94.80}{\scriptsize$\pm$0.09} & 80.74{\scriptsize$\pm$0.25}     & \textbf{90.33}{\scriptsize$\pm$0.54}     \\
            \bottomrule
        \end{tabu}
        }
    }% eof
    
    \caption{\textbf{POS tagging}, \textbf{dependency parsing} and {\bf NER} results on the test sets of our French datasets. \textit{UPOS (Universal Part-of-Speech) refers here to POS tagging accuracy, and LAS measures the overall accuracy of labeled dependencies in a parsed sentence.} }
    \label{tab:pos_and_dp_results}
\end{table*}

\paragraph{Question Answering.}
We evaluate our model on the FQuAD 1.0 dataset~\cite{2020arXiv200206071}, which is a SQuAD~\cite{rajpurkar-etal-2016-squad} style French question-answering dataset with 20731 examples for training, and 3188 for evaluation.

The results shown in Table~\ref{tab:qa} show that our model outperforms CamemBERT$_{30\%}$ by 6.01 F1 points, but shows no statistically significant improvement over CamemBERT$_{CCNet}$ F1 score, and exact match (EM) score.

\begin{table}[h]
    \centering
    {\footnotesize
        \begin{tabular}{lcc}
            \toprule
            Model               & F1             & EM             \\
            \midrule
            FrALBERT            & 72.6$^\ast$\textcolor{white}{XX0}    & 55.1$^\ast$\textcolor{white}{XXX}    \\
            \midrule
            CamemBERT$_{30\%}$  & 75.14{\scriptsize$\pm$0.17}          & 56.19{\scriptsize$\pm$0.27}          \\
            CamemBERT$_{CCNet}$ & \textbf{80.98}{\scriptsize$\pm$0.48}          & \textbf{62.51}{\scriptsize$\pm$0.54} \\
            
            \midrule
            \modelname          & \textbf{81.15}{\scriptsize$\pm$0.38} & \textbf{62.01}{\scriptsize$\pm$0.45}          \\
            \bottomrule
        \end{tabular}
    }%
    \caption{Question Answering results on FQuAD 1.0.}
    \label{tab:qa}
\end{table}

\paragraph{Part-of-Speech and Dependency Parsing.}
We report our results on 4 diverse French treebanks.
For the parser training, we make use of the HOPS parser~\cite{grobol:hal-03223424} implementation, which is a graph-based dependency parser inspired by \citet{dozat2017deep}.
Our configuration uses the Transformer model's last layer in addition to FastText embeddings~\cite{bojanowski-etal-2017-enriching}, character-level bi-directional RNN embeddings, and word embeddings trained during the fine-tuning phase.

Table~\ref{tab:pos_and_dp_results} shows that our proposed model consistently outperforms CamemBERT$_{30\%}$, and competes with CamemBERT$_{CCNet}$ on all 4 treebanks.

\paragraph{Named Entity Recognition} is performed on the French Treebank (FTB) which contains 350k tokens in 27k sentences extracted from news articles.
Our results in Table~\ref{tab:pos_and_dp_results},  surprisingly show that CamemBERT$_{30\%}$ outperforms CamemBERT$_{CCNet}$, while not being statistically better than our model.

\paragraph{FLUE Benchmark}
We use datasets from the French Language Understanding Evaluation (FLUE) benchmark~\cite{le-etal-2020-flaubert-unsupervised}, namely the French part of the paraphrase identification dataset PAWS-X~\cite{yang-etal-2019-paws}, and of XNLI~\cite{conneau-etal-2018-xnli}, in addition to CLS, a binary classification dataset with Amazon reviews taken from Amazon. 

Our results (Table~\ref{tab:text_classification_results}) show that our model outperforms all models on the CLS movie classification task, and matches the performance of CamemBERT$_{CCNet}$ on the other FLUE tasks.

\begin{table}[h]

    \small\centering
\resizebox{\columnwidth}{!}{%
    \begin{tabular}{lccc}
        \toprule
        Model               & \textsc{CLS} & \textsc{PAWS-X} & \textsc{XNLI}  \\
        \midrule
        FrALBERT  & 72.17{\scriptsize$\pm$3.32}           & 76.29{\scriptsize$\pm$1.28}           & 66.87{\scriptsize$\pm$0.42}      \\
        FlauBERT  & 93.22$^\ast$\textcolor{white}{000}            & 89.49$^\ast$\textcolor{white}{000}            & 80.6$^\ast$\textcolor{white}{0000}          \\
        \midrule
        CamemBERT$_{30\%}$  & 93.28{\scriptsize$\pm$0.19}            & 88.94{\scriptsize$\pm$0.14}          & 79.89{\scriptsize$\pm$0.64}          \\
        
        CamemBERT$_{CCNet}$ & 94.62{\scriptsize$\pm$0.04}            & \textbf{91.36}{\scriptsize$\pm$0.38}           & \textbf{81.95}{\scriptsize$\pm$0.51}          \\
        \midrule
        \modelname          & \textbf{94.92}{\scriptsize$\pm$0.13}   & \textbf{91.67}{\scriptsize$\pm$0.17}  & \textbf{82.00}{\scriptsize$\pm$0.17} \\
        \bottomrule
    \end{tabular}%
    }
    \caption{Text classification results (Accuracy) on the FLUE benchmark. \textit{$^\ast$Results taken from \citet{le-etal-2020-flaubert-unsupervised}}.}
    \label{tab:text_classification_results}
\end{table}

\paragraph{Pre-training Dataset Choice}
\label{app:ccnet}
We choose CCNet as our pre-training dataset instead of the more common OSCAR dataset~\cite{OrtizSuarezSagotRomary2019}, as (i) it was shown to produce less offensive output~\cite{launay-etal-2022-pagnol} and (ii) it allowed us to be fully comparable with many of the CamemBERT models \cite{martin-etal-2020-camembert}, enabling thus meaningful comparisons.
Nevertheless, we also ran experiments with CamemBERT$_{OSCAR}$, and found that it performed slightly worse than CamemBERT$_{CCNet}$, as shown in Table~\ref{tab:negative-results} Appendix~\ref{app:appendix}.

\paragraph{Pre-training Compute and CO2 Impact}
Our model was trained for 8 days on 6 A40 GPUs, compared to CamemBERT which was trained on 256 V100 GPUs for one day, which is roughly equivalent to 28 days of training on 6 A40 GPUs, since an NVIDIA A40 GPU is about 1.5x faster than a V100 GPU on language modeling tasks according to recent benchmarks.\footnote{See \href{https://lambdalabs.com/blog/nvidia-rtx-a40-benchmarks}{https://lambdalabs.com/blog/nvidia-rtx-a40-benchmarks}.}

Following the reports by \citet{luccioni2022estimating} and \citet{cattan2022benchmarking} on the environmental impact of language model training, we use \citeauthor{lannelongue2021green}'s \citeyearpar{lannelongue2021green} online carbon footprint calculator to provide the following estimates: \modelname's pre-training used 700kWh and emitted 36kg CO$_2$ compared to 3.32MWh and 170kg for CamemBERT.\footnote{These estimates are specific to our training infrastructure situated in France.
These estimates highlight the remarkable efficiency achieved by CamemBERTa's pretraining process.
}

%% file: sections/discussion.tex
\section{Discussion}
Our experiments clearly show that given the same training corpus, tokenizer, and total number of examples seen during training, \modelname outperforms the MLM trained CamemBERT model on all tasks except NER on FTB and POS tagging on Rhapsodie. Moreover, our model implementation is able to match or outperform a fully trained CamemBERT model, trained on around 3 times more samples and more compute. The strong performance of our model on higher level FLUE tasks suggest that lower level tasks such as POS tagging and dependency parsing are less challenging for current generation models, since they mostly require surface level information which the model can capture early in the training process, as suggested by \citet{martin-etal-2020-camembert}, compared to tasks such as question answering and text classification which require more complex processing.

Taking a step back and looking at the only DeBERTa model that includes French,  mDeBERTa  \cite{he2021debertav3} we can see (cf. Table \ref{tab:mdeberta}) that our model only requires 6.6\% of its multilingual counterpart training samples to achieve competitive performance while additionally also outperforming the XLM-R model  \cite{conneau-etal-2020-unsupervised} trained on a  much larger training sample size.

\begin{table}[h!]
\begin{center}
{\footnotesize
\resizebox{\columnwidth}{!}{
\begin{tabular}{lcccc}
\hline
{ } & { XNLI} & { Steps} & { \# tokens$^\dagger$} & {Size$^\ddagger$} \\
\toprule
mDeBERTa$^*$ & 84.4 & 500k & 2T & 0.295T \\
\modelname & 82.0 & 33k$^{\dagger\dagger}$ & 0.139T & 0.032T \\
\midrule
XLM-R$^{**}$ & 81.4 & 1.5M & 6T & 0.295T \\
{ C.BERT$_{CCNet}$} & { 81.95} & { 100k} & { 0.419T} & { 0.032T} \\
\bottomrule
\end{tabular}
}
\caption{Comparison of XNLI results for different pre-training settings. $^{\dagger\dagger}$step count was converted assuming 8k batch size. $^\dagger$the total number of tokens seen during training. $^\ddagger$Total dataset size in tokens. $^*$\citet{he2021debertav3}, $^{**}$\citet{conneau-etal-2020-unsupervised}.}
\label{tab:mdeberta}
}%eof foot
\end{center}
\end{table}

This confirms the interest in using such training paradigms in compute limited scenarios for semantically demanding tasks such as question-answering or natural-language inference.

Last but not least, other competitive language models for French are available and although  not the primary focus of this paper, we conducted a comparative analysis involving FlauBERT~\cite{le-etal-2020-flaubert-unsupervised} and FrALBERT~\cite{cattan:hal-03336060}. 
The results, presented in Table~\ref{tab:negative-results} in Appendix~\ref{app:appendix}, demonstrate the better performance of  our model across all evaluated tasks in comparison to these French models. Additionally, it is worth noting that FlauBERT was trained for 17 days with 32 V100 GPUs, which is equivalent to 60 days of training on 6 A40 GPUs.  This represents a 7.5-fold increase in computational resources employed compared to \modelname.

%% file: sections/conclusion.tex
\section{Conclusion}
We presented \modelname, a data-efficient French language model trained on a large corpus of French text and the first publicly available DeBERTaV3-style pretrained model and implementation.
For a fair evaluation we reused the same corpus and tokenizer as CamemBERT$_{CCNet}$, but using only 30\% of the total number of input training tokens.
We compared the performance of both models in addition to an MLM model trained from scratch under the same setup as \modelname, CamemBERT$_{30\%}$, on a variety of downstream tasks.
Our experiments showed that our model outperforms CamemBERT$_{30\%}$ on all tasks except NER on FTB, and that it is able to match and even surpass CamemBERT$_{CCNet}$.
Furthermore, we have also made our optimized code implementation and pretrained model weights publicly available for others to use.

%% file: sections/extra.tex
\section*{Limitations}
Although our model is more efficient than previous models trained using the MLM objective and the standard transformer architecture, we notice that the models runs around 30\% slower.
This is due to the disentangled attention mechanism, which is more computationally expensive than the standard attention mechanism.
We also note that at the time of writing, the DeBERTaV3 TensorFLow 2 implementation available on HuggingFace's Transformers library~\cite{wolf-etal-2020-transformers} experiences heavy slowdowns with TPU backends.
Our attempts to solve this issue were unsuccessful, and we were unable to train our model on TPUs.

\section*{Ethics Statement}
We propose a model trained using DeBERTaV3 style pre-training along with an optimized training implementation, which reduces training computation cost when compared to previous models, and hence greatly reduces the energy cost and environmental impact of language model training.
We trained our model using the CCNet dataset, for which we direct the reader to for further discussion on bias and ethical considerations.
Our experiments do not include any additional data collection or human annotators.
Like other language models trained on massive corpora, there may be potential biases present in the training data, which could affect the output of our models. 
Therefore, we advise against using these models in production without thorough testing.
All our experiments were carried out on clusters with energy sources consisting of nuclear (65--75\%), 20\% renewable, and the remaining from gas.

\section*{Acknowledgements}

This work was partly funded by Benoît Sagot's chair in the PRAIRIE institute funded by the French national reseach agency (ANR as part of the ``Investissements d’avenir'' programme under the reference \mbox{ANR-19-P3IA-0001}. This work  also received
funding from the European Union’s Horizon 2020
research and innovation programme under grant
agreement No. 101021607. The authors are grateful to the OPAL infrastructure from Université Côte d'Azur for providing resources and support.

%% file: sections/appendix.tex
\section*{Appendix}

\section{Experiments Results on OSCAR and Dropout}
\label{app:appendix}

\begin{table*}[h!]
  \centering
  {\footnotesize
    \begin{tabular}{lcccccccc}
      \toprule
      Model                  & UPOS  & LAS   & NER   & CLS & PAWS-X & XNLI  & F1$_{FQuAD}$ & EM$_{FQuAD}$ \\
      \midrule
      FrALBERT    & 93.53 & 78.89 & 69.83 & 72.17   & 76.29  & 66.87 & 72.6$^\ast$       & 55.1$^\ast$       \\
      FlauBERT    & 97.51 & 87.92 & -   & 93.22$^\ast$    & 89.49$^\ast$        & 80.6$^\ast$   & -   & -      \\
      \midrule
      CamemBERT$_{OSCAR}$    & 97.50 & 88.24 & 88.19 & 94.61   & 90.87  & 81.38 & 79.92        & 61.15        \\
      CamemBERT$_{CCNet}$    & \underline{97.59} & \underline{88.69} & \underline{89.97} & \underline{94.62}   & \underline{91.36}  & \underline{81.95} & \underline{80.98}        & \underline{62.51}        \\
      \midrule
      \modelname             & \underline{97.57} & 88.55 & \underline{90.33} & \underline{94.92}   & \underline{91.67}  & \underline{82.00} & \underline{81.15}        & \underline{62.01}        \\
      \textsc{CamemBERTa}$_{dropout}$ & 97.56 & \underline{88.57} & 90.03 & 94.46   & 91.42  & 81.91 & 79.37        & 60.29        \\
      \bottomrule
    \end{tabular}
  }
  \caption{Comparison results of CamemBERT$_{OSCAR}$ and CamemBERT$_{CCNet}$, and our model \modelname, with and without dropout. Due to compatibility issues with FlauBERT's tokenizer, we were unable to conduct FlauBERT testing on FQuAD and NER using standard finetuning scripts. \textit{$^\ast$Results from the models' respective papers \citet{cattan:hal-03336060} and ~\cite{le-etal-2020-flaubert-unsupervised}}.}
  \label{tab:negative-results}
\end{table*}

\section{Negative Results}
\label{sec:neg-res-appendix}
In addition to our main results, we attempted to improve the performance of our model by adding BPE-Dropout~\cite{provilkov-etal-2020-bpe} to the tokenization process, as it was shown that this method of subword regularization improves performance on translation tasks.
We retrain our model with BPE-Dropout, dubbed CamemBERTa$_{dropout}$, and compare the results to our original model in Table~\ref{tab:negative-results}.
We observe that by adding BPE-Dropout, we obtain a decrease in performance on most tasks, except for POS tagging and dependency parsing, where the performance does not change.

\section{Hyper-parameters}
\label{sec:HP-appendix}

\begin{table}[h!]
    \centering
    \begin{tabular}{lc}
        \toprule
        Hyper-parameter     & Value                \\
        \midrule
        Max sequence length & 512                  \\
        Batch size          & 16                   \\
        FP16                & Enabled              \\
        Learning rate       & \{1.5e-5,2e-5,3e-5\} \\
        Epochs              & 8                    \\
        Scheduler           & linear               \\
        Warmup steps        & \{0,0.1\%\}          \\
        Seed                & \{1,25,42,666,1337\} \\
        \bottomrule
    \end{tabular}
    \caption{Hyper-parameters used for the Question Answering and Named Entity Recognition experiments.}
    \label{tab:hyperparams}
\end{table}

For experiments on the FLUE benchmark we use the same hyper-parameters as the authors of CamemBERT on the NLI task.
As for POS tagging and dependency parsing, we use the same configurations as the one used in \citet{riabi-etal-2021-character}.

%% file: acl2023.bbl
\begin{thebibliography}{53}
\expandafter\ifx\csname natexlab\endcsname\relax\def\natexlab#1{#1}\fi

\bibitem[{Abeill{\'e} et~al.(2000)Abeill{\'e}, Cl{\'e}ment, and
  Kinyon}]{abeille-etal-2000-building}
Anne Abeill{\'e}, Lionel Cl{\'e}ment, and Alexandra Kinyon. 2000.
\newblock \href {http://www.lrec-conf.org/proceedings/lrec2000/pdf/230.pdf}
  {Building a treebank for {F}rench}.
\newblock In \emph{Proceedings of the Second International Conference on
  Language Resources and Evaluation ({LREC}{'}00)}, Athens, Greece. European
  Language Resources Association (ELRA).

\bibitem[{Bojanowski et~al.(2017)Bojanowski, Grave, Joulin, and
  Mikolov}]{bojanowski-etal-2017-enriching}
Piotr Bojanowski, Edouard Grave, Armand Joulin, and Tomas Mikolov. 2017.
\newblock \href {https://doi.org/10.1162/tacl_a_00051} {Enriching word vectors
  with subword information}.
\newblock \emph{Transactions of the Association for Computational Linguistics},
  5:135--146.

\bibitem[{Candito and Crabb{\'e}(2009)}]{candito-crabbe-2009-improving}
Marie Candito and Beno{\^\i}t Crabb{\'e}. 2009.
\newblock \href {https://aclanthology.org/W09-3821} {Improving generative
  statistical parsing with semi-supervised word clustering}.
\newblock In \emph{Proceedings of the 11th International Conference on Parsing
  Technologies ({IWPT}{'}09)}, pages 138--141, Paris, France. Association for
  Computational Linguistics.

\bibitem[{Candito et~al.(2014)Candito, Perrier, Guillaume, Ribeyre, Fort,
  Seddah, and Clergerie}]{CANDITO14.494}
Marie Candito, Guy Perrier, Bruno Guillaume, Corentin Ribeyre, Karën Fort,
  Djamé Seddah, and Eric De~La Clergerie. 2014.
\newblock Deep syntax annotation of the sequoia french treebank.
\newblock In \emph{Proceedings of the Ninth International Conference on
  Language Resources and Evaluation (LREC'14)}, Reykjavik, Iceland. European
  Language Resources Association (ELRA).

\bibitem[{Candito and Seddah(2012)}]{candito-seddah-2012-le}
Marie Candito and Djam{\'e} Seddah. 2012.
\newblock \href {https://aclanthology.org/F12-2024} {Le corpus sequoia :
  annotation syntaxique et exploitation pour l{'}adaptation d{'}analyseur par
  pont lexical (the sequoia corpus : Syntactic annotation and use for a parser
  lexical domain adaptation method) [in {F}rench]}.
\newblock In \emph{Proceedings of the Joint Conference JEP-TALN-RECITAL 2012,
  volume 2: TALN}, pages 321--334, Grenoble, France. ATALA/AFCP.

\bibitem[{Cattan et~al.(2022)Cattan, Ghannay, Servan, and
  Rosset}]{cattan2022benchmarking}
Oralie Cattan, Sahar Ghannay, Christophe Servan, and Sophie Rosset. 2022.
\newblock Benchmarking transformers-based models on french spoken language
  understanding tasks.
\newblock \emph{arXiv preprint arXiv:2207.09152}.

\bibitem[{Cattan et~al.(2021)Cattan, Servan, and Rosset}]{cattan:hal-03336060}
Oralie Cattan, Christophe Servan, and Sophie Rosset. 2021.
\newblock \href {https://hal.science/hal-03336060} {{On the Usability of
  Transformers-based models for a French Question-Answering task}}.
\newblock In \emph{{Recent Advances in Natural Language Processing (RANLP)}},
  Varna, Bulgaria.

\bibitem[{Clark et~al.(2020)Clark, Luong, Le, and Manning}]{clark2020electra}
Kevin Clark, Minh-Thang Luong, Quoc~V. Le, and Christopher~D. Manning. 2020.
\newblock \href {https://openreview.net/pdf?id=r1xMH1BtvB} {{ELECTRA}:
  Pre-training text encoders as discriminators rather than generators}.
\newblock In \emph{ICLR}.

\bibitem[{Conneau et~al.(2020)Conneau, Khandelwal, Goyal, Chaudhary, Wenzek,
  Guzm{\'a}n, Grave, Ott, Zettlemoyer, and
  Stoyanov}]{conneau-etal-2020-unsupervised}
Alexis Conneau, Kartikay Khandelwal, Naman Goyal, Vishrav Chaudhary, Guillaume
  Wenzek, Francisco Guzm{\'a}n, Edouard Grave, Myle Ott, Luke Zettlemoyer, and
  Veselin Stoyanov. 2020.
\newblock \href {https://doi.org/10.18653/v1/2020.acl-main.747} {Unsupervised
  cross-lingual representation learning at scale}.
\newblock In \emph{Proceedings of the 58th Annual Meeting of the Association
  for Computational Linguistics}, pages 8440--8451, Online. Association for
  Computational Linguistics.

\bibitem[{Conneau et~al.(2018)Conneau, Rinott, Lample, Williams, Bowman,
  Schwenk, and Stoyanov}]{conneau-etal-2018-xnli}
Alexis Conneau, Ruty Rinott, Guillaume Lample, Adina Williams, Samuel Bowman,
  Holger Schwenk, and Veselin Stoyanov. 2018.
\newblock \href {https://doi.org/10.18653/v1/D18-1269} {{XNLI}: Evaluating
  cross-lingual sentence representations}.
\newblock In \emph{Proceedings of the 2018 Conference on Empirical Methods in
  Natural Language Processing}, pages 2475--2485, Brussels, Belgium.
  Association for Computational Linguistics.

\bibitem[{Dai et~al.(2019)Dai, Yang, Yang, Carbonell, Le, and
  Salakhutdinov}]{dai-etal-2019-transformer}
Zihang Dai, Zhilin Yang, Yiming Yang, Jaime Carbonell, Quoc Le, and Ruslan
  Salakhutdinov. 2019.
\newblock \href {https://doi.org/10.18653/v1/P19-1285} {Transformer-{XL}:
  Attentive language models beyond a fixed-length context}.
\newblock In \emph{Proceedings of the 57th Annual Meeting of the Association
  for Computational Linguistics}, pages 2978--2988, Florence, Italy.
  Association for Computational Linguistics.

\bibitem[{d'Hoffschmidt et~al.(2020)d'Hoffschmidt, Vidal, Belblidia, and
  Brendlé}]{2020arXiv200206071}
Martin d'Hoffschmidt, Maxime Vidal, Wacim Belblidia, and Tom Brendlé. 2020.
\newblock \href {http://arxiv.org/abs/2002.06071} {{FQuAD: French Question
  Answering Dataset}}.
\newblock \emph{arXiv e-prints}, page arXiv:2002.06071.

\bibitem[{Dozat and Manning(2017)}]{dozat2017deep}
Timothy Dozat and Christopher~D. Manning. 2017.
\newblock \href {https://openreview.net/forum?id=Hk95PK9le} {Deep biaffine
  attention for neural dependency parsing}.
\newblock In \emph{International Conference on Learning Representations}.

\bibitem[{Fedus et~al.(2021)Fedus, Zoph, and Shazeer}]{fedus2021switch}
William Fedus, Barret Zoph, and Noam Shazeer. 2021.
\newblock Switch transformers: Scaling to trillion parameter models with simple
  and efficient sparsity.
\newblock \emph{arXiv preprint arXiv:2101.03961}.

\bibitem[{Gabay et~al.(2022)Gabay, Ortiz~Suarez, Bartz, Chagu{\'e}, Bawden,
  Gambette, and Sagot}]{gabay-etal-2022-freem}
Simon Gabay, Pedro Ortiz~Suarez, Alexandre Bartz, Alix Chagu{\'e}, Rachel
  Bawden, Philippe Gambette, and Beno{\^\i}t Sagot. 2022.
\newblock \href {https://aclanthology.org/2022.lrec-1.359} {From {F}re{EM} to
  d{'}{A}lem{BERT}: a large corpus and a language model for early {M}odern
  {F}rench}.
\newblock In \emph{Proceedings of the Thirteenth Language Resources and
  Evaluation Conference}, pages 3367--3374, Marseille, France. European
  Language Resources Association.

\bibitem[{Geiping and Goldstein(2022)}]{geiping2022cramming}
Jonas Geiping and Tom Goldstein. 2022.
\newblock Cramming: Training a language model on a single gpu in one day.

\bibitem[{Grobol and Crabbé(2021)}]{grobol:hal-03223424}
Loïc Grobol and Benoît Crabbé. 2021.
\newblock \href {https://hal.archives-ouvertes.fr/hal-03223424} {{Analyse en
  dépendances du français avec des plongements contextualisés}}.
\newblock In \emph{{Actes de la 28ème Conférence sur le Traitement
  Automatique des Langues Naturelles}}.

\bibitem[{He et~al.(2021{\natexlab{a}})He, Gao, and Chen}]{he2021debertav3}
Pengcheng He, Jianfeng Gao, and Weizhu Chen. 2021{\natexlab{a}}.
\newblock \href {http://arxiv.org/abs/2111.09543} {Debertav3: Improving deberta
  using electra-style pre-training with gradient-disentangled embedding
  sharing}.

\bibitem[{He et~al.(2021{\natexlab{b}})He, Liu, Gao, and Chen}]{he2021deberta}
Pengcheng He, Xiaodong Liu, Jianfeng Gao, and Weizhu Chen. 2021{\natexlab{b}}.
\newblock \href {https://openreview.net/forum?id=XPZIaotutsD} {Deberta:
  Decoding-enhanced bert with disentangled attention}.
\newblock In \emph{International Conference on Learning Representations}.

\bibitem[{Hoffmann et~al.(2022)Hoffmann, Borgeaud, Mensch, Buchatskaya, Cai,
  Rutherford, Casas, Hendricks, Welbl, Clark et~al.}]{hoffmann2022training}
Jordan Hoffmann, Sebastian Borgeaud, Arthur Mensch, Elena Buchatskaya, Trevor
  Cai, Eliza Rutherford, Diego de~Las Casas, Lisa~Anne Hendricks, Johannes
  Welbl, Aidan Clark, et~al. 2022.
\newblock Training compute-optimal large language models.
\newblock \emph{arXiv preprint arXiv:2203.15556}.

\bibitem[{Izsak et~al.(2021)Izsak, Berchansky, and
  Levy}]{izsak-etal-2021-train}
Peter Izsak, Moshe Berchansky, and Omer Levy. 2021.
\newblock \href {https://doi.org/10.18653/v1/2021.emnlp-main.831} {How to train
  {BERT} with an academic budget}.
\newblock In \emph{Proceedings of the 2021 Conference on Empirical Methods in
  Natural Language Processing}, pages 10644--10652, Online and Punta Cana,
  Dominican Republic. Association for Computational Linguistics.

\bibitem[{Kamal~Eddine et~al.(2021)Kamal~Eddine, Tixier, and
  Vazirgiannis}]{kamal-eddine-etal-2021-barthez}
Moussa Kamal~Eddine, Antoine Tixier, and Michalis Vazirgiannis. 2021.
\newblock \href {https://doi.org/10.18653/v1/2021.emnlp-main.740} {{BART}hez: a
  skilled pretrained {F}rench sequence-to-sequence model}.
\newblock In \emph{Proceedings of the 2021 Conference on Empirical Methods in
  Natural Language Processing}, pages 9369--9390, Online and Punta Cana,
  Dominican Republic. Association for Computational Linguistics.

\bibitem[{Kudo and Richardson(2018)}]{kudo2018sentencepiece}
Taku Kudo and John Richardson. 2018.
\newblock \href {https://www.aclweb.org/anthology/D18-2012/} {Sentencepiece:
  {A} simple and language independent subword tokenizer and detokenizer for
  neural text processing}.
\newblock In \emph{Proceedings of the 2018 Conference on Empirical Methods in
  Natural Language Processing, {EMNLP} 2018: System Demonstrations, Brussels,
  Belgium, October 31 - November 4, 2018}, pages 66--71.

\bibitem[{Lacheret et~al.(2014)Lacheret, Kahane, Beliao, Dister, Gerdes,
  Goldman, Obin, Pietrandrea, and Tchobanov}]{lacheret:halshs-01061368}
Anne Lacheret, Sylvain Kahane, Julie Beliao, Anne Dister, Kim Gerdes,
  Jean-Philippe Goldman, Nicolas Obin, Paola Pietrandrea, and Atanas Tchobanov.
  2014.
\newblock \href {https://doi.org/10.1051/shsconf/20140801305} {{Rhapsodie: un
  Treebank annot{\'e} pour l'{\'e}tude de l'interface syntaxe-prosodie en
  fran{\c c}ais parl{\'e}}}.
\newblock In \emph{{4e Congr{\`e}s Mondial de Linguistique Fran{\c c}aise}},
  volume~8, pages 2675--2689, Berlin, Germany.

\bibitem[{Lan et~al.(2020)Lan, Chen, Goodman, Gimpel, Sharma, and
  Soricut}]{Lan2020ALBERT}
Zhenzhong Lan, Mingda Chen, Sebastian Goodman, Kevin Gimpel, Piyush Sharma, and
  Radu Soricut. 2020.
\newblock \href {https://openreview.net/forum?id=H1eA7AEtvS} {Albert: A lite
  bert for self-supervised learning of language representations}.
\newblock In \emph{International Conference on Learning Representations}.

\bibitem[{Lannelongue et~al.(2021)Lannelongue, Grealey, and
  Inouye}]{lannelongue2021green}
Lo{\"\i}c Lannelongue, Jason Grealey, and Michael Inouye. 2021.
\newblock Green algorithms: quantifying the carbon footprint of computation.
\newblock \emph{Advanced science}, 8(12):2100707.

\bibitem[{Launay et~al.(2022)Launay, Tommasone, Pannier, Boniface, Chatelain,
  Cappelli, Poli, and Seddah}]{launay-etal-2022-pagnol}
Julien Launay, E.l. Tommasone, Baptiste Pannier, Fran{\c{c}}ois Boniface,
  Am{\'e}lie Chatelain, Alessandro Cappelli, Iacopo Poli, and Djam{\'e} Seddah.
  2022.
\newblock \href {https://aclanthology.org/2022.lrec-1.455} {{PAG}nol: An
  extra-large {F}rench generative model}.
\newblock In \emph{Proceedings of the Thirteenth Language Resources and
  Evaluation Conference}, pages 4275--4284, Marseille, France. European
  Language Resources Association.

\bibitem[{Le et~al.(2020)Le, Vial, Frej, Segonne, Coavoux, Lecouteux, Allauzen,
  Crabb{\'e}, Besacier, and Schwab}]{le-etal-2020-flaubert-unsupervised}
Hang Le, Lo{\"\i}c Vial, Jibril Frej, Vincent Segonne, Maximin Coavoux,
  Benjamin Lecouteux, Alexandre Allauzen, Benoit Crabb{\'e}, Laurent Besacier,
  and Didier Schwab. 2020.
\newblock \href {https://aclanthology.org/2020.lrec-1.302} {{F}lau{BERT}:
  Unsupervised language model pre-training for {F}rench}.
\newblock In \emph{Proceedings of the Twelfth Language Resources and Evaluation
  Conference}, pages 2479--2490, Marseille, France. European Language Resources
  Association.

\bibitem[{Lewis et~al.(2020)Lewis, Liu, Goyal, Ghazvininejad, Mohamed, Levy,
  Stoyanov, and Zettlemoyer}]{lewis-etal-2020-bart}
Mike Lewis, Yinhan Liu, Naman Goyal, Marjan Ghazvininejad, Abdelrahman Mohamed,
  Omer Levy, Veselin Stoyanov, and Luke Zettlemoyer. 2020.
\newblock \href {https://doi.org/10.18653/v1/2020.acl-main.703} {{BART}:
  Denoising sequence-to-sequence pre-training for natural language generation,
  translation, and comprehension}.
\newblock In \emph{Proceedings of the 58th Annual Meeting of the Association
  for Computational Linguistics}, pages 7871--7880, Online. Association for
  Computational Linguistics.

\bibitem[{Liu et~al.(2020)Liu, Ott, Goyal, Du, Joshi, Chen, Levy, Lewis,
  Zettlemoyer, and Stoyanov}]{liu2020roberta}
Yinhan Liu, Myle Ott, Naman Goyal, Jingfei Du, Mandar Joshi, Danqi Chen, Omer
  Levy, Mike Lewis, Luke Zettlemoyer, and Veselin Stoyanov. 2020.
\newblock \href {https://openreview.net/forum?id=SyxS0T4tvS} {Ro{\{}bert{\}}a:
  A robustly optimized {\{}bert{\}} pretraining approach}.

\bibitem[{Luccioni et~al.(2022)Luccioni, Viguier, and
  Ligozat}]{luccioni2022estimating}
Alexandra~Sasha Luccioni, Sylvain Viguier, and Anne-Laure Ligozat. 2022.
\newblock Estimating the carbon footprint of bloom, a 176b parameter language
  model.
\newblock \emph{arXiv preprint arXiv:2211.02001}.

\bibitem[{Martin et~al.(2020)Martin, Muller, Ortiz~Su{\'a}rez, Dupont, Romary,
  de~la Clergerie, Seddah, and Sagot}]{martin-etal-2020-camembert}
Louis Martin, Benjamin Muller, Pedro~Javier Ortiz~Su{\'a}rez, Yoann Dupont,
  Laurent Romary, {\'E}ric de~la Clergerie, Djam{\'e} Seddah, and Beno{\^\i}t
  Sagot. 2020.
\newblock \href {https://doi.org/10.18653/v1/2020.acl-main.645} {{C}amem{BERT}:
  a tasty {F}rench language model}.
\newblock In \emph{Proceedings of the 58th Annual Meeting of the Association
  for Computational Linguistics}, pages 7203--7219, Online. Association for
  Computational Linguistics.

\bibitem[{McDonald et~al.(2013)McDonald, Nivre, Quirmbach-Brundage, Goldberg,
  Das, Ganchev, Hall, Petrov, Zhang, T{\"a}ckstr{\"o}m, Bedini,
  Bertomeu~Castell{\'o}, and Lee}]{mcdonald-etal-2013-universal}
Ryan McDonald, Joakim Nivre, Yvonne Quirmbach-Brundage, Yoav Goldberg, Dipanjan
  Das, Kuzman Ganchev, Keith Hall, Slav Petrov, Hao Zhang, Oscar
  T{\"a}ckstr{\"o}m, Claudia Bedini, N{\'u}ria Bertomeu~Castell{\'o}, and
  Jungmee Lee. 2013.
\newblock \href {https://aclanthology.org/P13-2017} {{U}niversal {D}ependency
  annotation for multilingual parsing}.
\newblock In \emph{Proceedings of the 51st Annual Meeting of the Association
  for Computational Linguistics (Volume 2: Short Papers)}, pages 92--97, Sofia,
  Bulgaria. Association for Computational Linguistics.

\bibitem[{Micheli et~al.(2020)Micheli, d{'}Hoffschmidt, and
  Fleuret}]{micheli-etal-2020-importance}
Vincent Micheli, Martin d{'}Hoffschmidt, and Fran{\c{c}}ois Fleuret. 2020.
\newblock \href {https://doi.org/10.18653/v1/2020.emnlp-main.632} {On the
  importance of pre-training data volume for compact language models}.
\newblock In \emph{Proceedings of the 2020 Conference on Empirical Methods in
  Natural Language Processing (EMNLP)}, pages 7853--7858, Online. Association
  for Computational Linguistics.

\bibitem[{M{\"{u}}ller and Laurent(2022)}]{muller2022cedille}
Martin M{\"{u}}ller and Florian Laurent. 2022.
\newblock \href {http://arxiv.org/abs/2202.03371} {Cedille: A large
  autoregressive french language model}.

\bibitem[{{Ortiz Su{\'a}rez} et~al.(2019){Ortiz Su{\'a}rez}, Sagot, and
  Romary}]{OrtizSuarezSagotRomary2019}
Pedro~Javier {Ortiz Su{\'a}rez}, Beno{\^i}t Sagot, and Laurent Romary. 2019.
\newblock \href {https://doi.org/10.14618/ids-pub-9021} {Asynchronous pipelines
  for processing huge corpora on medium to low resource infrastructures}.
\newblock Proceedings of the Workshop on Challenges in the Management of Large
  Corpora (CMLC-7) 2019. Cardiff, 22nd July 2019, pages 9 -- 16, Mannheim.
  Leibniz-Institut f{\"u}r Deutsche Sprache.

\bibitem[{Pan et~al.(2022)Pan, Diao, Chen, and Zhang}]{extreme-bert}
Rui Pan, Shizhe Diao, Jianlin Chen, and Tong Zhang. 2022.
\newblock \href {http://arxiv.org/abs/2211.17201} {Extremebert: A toolkit for
  accelerating pretraining of customized bert}.

\bibitem[{Provilkov et~al.(2020)Provilkov, Emelianenko, and
  Voita}]{provilkov-etal-2020-bpe}
Ivan Provilkov, Dmitrii Emelianenko, and Elena Voita. 2020.
\newblock \href {https://doi.org/10.18653/v1/2020.acl-main.170} {{BPE}-dropout:
  Simple and effective subword regularization}.
\newblock In \emph{Proceedings of the 58th Annual Meeting of the Association
  for Computational Linguistics}, pages 1882--1892, Online. Association for
  Computational Linguistics.

\bibitem[{Qu et~al.(2021)Qu, Niu, and Mo}]{qu-etal-2021-explore}
Anlin Qu, Jianwei Niu, and Shasha Mo. 2021.
\newblock \href {https://doi.org/10.18653/v1/2021.emnlp-main.237} {Explore
  better relative position embeddings from encoding perspective for transformer
  models}.
\newblock In \emph{Proceedings of the 2021 Conference on Empirical Methods in
  Natural Language Processing}, pages 2989--2997, Online and Punta Cana,
  Dominican Republic. Association for Computational Linguistics.

\bibitem[{Radford et~al.(2019)Radford, Wu, Child, Luan, Amodei, and
  Sutskever}]{radford2019language}
Alec Radford, Jeff Wu, Rewon Child, David Luan, Dario Amodei, and Ilya
  Sutskever. 2019.
\newblock Language models are unsupervised multitask learners.

\bibitem[{Rae et~al.(2021)Rae, Borgeaud, Cai, Millican, Hoffmann, Song,
  Aslanides, Henderson, Ring, Young et~al.}]{rae2021scaling}
Jack~W Rae, Sebastian Borgeaud, Trevor Cai, Katie Millican, Jordan Hoffmann,
  Francis Song, John Aslanides, Sarah Henderson, Roman Ring, Susannah Young,
  et~al. 2021.
\newblock Scaling language models: Methods, analysis \& insights from training
  gopher.
\newblock \emph{arXiv preprint arXiv:2112.11446}.

\bibitem[{Raffel et~al.(2020)Raffel, Shazeer, Roberts, Lee, Narang, Matena,
  Zhou, Li, and Liu}]{raffel2019exploring}
Colin Raffel, Noam Shazeer, Adam Roberts, Katherine Lee, Sharan Narang, Michael
  Matena, Yanqi Zhou, Wei Li, and Peter~J. Liu. 2020.
\newblock \href {http://jmlr.org/papers/v21/20-074.html} {Exploring the limits
  of transfer learning with a unified text-to-text transformer}.
\newblock \emph{Journal of Machine Learning Research}, 21(140):1--67.

\bibitem[{Rajpurkar et~al.(2016)Rajpurkar, Zhang, Lopyrev, and
  Liang}]{rajpurkar-etal-2016-squad}
Pranav Rajpurkar, Jian Zhang, Konstantin Lopyrev, and Percy Liang. 2016.
\newblock \href {https://doi.org/10.18653/v1/D16-1264} {{SQ}u{AD}: 100,000+
  questions for machine comprehension of text}.
\newblock In \emph{Proceedings of the 2016 Conference on Empirical Methods in
  Natural Language Processing}, pages 2383--2392, Austin, Texas. Association
  for Computational Linguistics.

\bibitem[{Riabi et~al.(2021)Riabi, Sagot, and
  Seddah}]{riabi-etal-2021-character}
Arij Riabi, Beno{\^\i}t Sagot, and Djam{\'e} Seddah. 2021.
\newblock \href {https://doi.org/10.18653/v1/2021.wnut-1.47} {Can
  character-based language models improve downstream task performances in
  low-resource and noisy language scenarios?}
\newblock In \emph{Proceedings of the Seventh Workshop on Noisy User-generated
  Text (W-NUT 2021)}, pages 423--436, Online. Association for Computational
  Linguistics.

\bibitem[{Sagot et~al.(2012)Sagot, Richard, and
  Stern}]{sagot-etal-2012-annotation}
Beno{\^\i}t Sagot, Marion Richard, and Rosa Stern. 2012.
\newblock \href {https://aclanthology.org/F12-2050} {Annotation
  r{\'e}f{\'e}rentielle du corpus arbor{\'e} de {P}aris 7 en entit{\'e}s
  nomm{\'e}es (referential named entity annotation of the {P}aris 7 {F}rench
  {T}ree{B}ank) [in {F}rench]}.
\newblock In \emph{Proceedings of the Joint Conference JEP-TALN-RECITAL 2012,
  volume 2: TALN}, pages 535--542, Grenoble, France. ATALA/AFCP.

\bibitem[{Seddah et~al.(2012)Seddah, Sagot, Candito, Mouilleron, and
  Combet}]{seddah-etal-2012-french}
Djam{\'e} Seddah, Benoit Sagot, Marie Candito, Virginie Mouilleron, and Vanessa
  Combet. 2012.
\newblock \href {https://aclanthology.org/C12-1149} {The {F}rench {S}ocial
  {M}edia {B}ank: a treebank of noisy user generated content}.
\newblock In \emph{Proceedings of {COLING} 2012}, pages 2441--2458, Mumbai,
  India. The COLING 2012 Organizing Committee.

\bibitem[{Sergeev and Balso(2018)}]{sergeev2018horovod}
Alexander Sergeev and Mike~Del Balso. 2018.
\newblock Horovod: fast and easy distributed deep learning in {TensorFlow}.
\newblock \emph{arXiv preprint arXiv:1802.05799}.

\bibitem[{Shaw et~al.(2018)Shaw, Uszkoreit, and Vaswani}]{shaw-etal-2018-self}
Peter Shaw, Jakob Uszkoreit, and Ashish Vaswani. 2018.
\newblock \href {https://doi.org/10.18653/v1/N18-2074} {Self-attention with
  relative position representations}.
\newblock In \emph{Proceedings of the 2018 Conference of the North {A}merican
  Chapter of the Association for Computational Linguistics: Human Language
  Technologies, Volume 2 (Short Papers)}, pages 464--468, New Orleans,
  Louisiana. Association for Computational Linguistics.

\bibitem[{Vaswani et~al.(2017)Vaswani, Shazeer, Parmar, Uszkoreit, Jones,
  Gomez, Kaiser, and Polosukhin}]{vaswani2017attention}
Ashish Vaswani, Noam Shazeer, Niki Parmar, Jakob Uszkoreit, Llion Jones,
  Aidan~N Gomez, \L~ukasz Kaiser, and Illia Polosukhin. 2017.
\newblock \href
  {https://proceedings.neurips.cc/paper/2017/file/3f5ee243547dee91fbd053c1c4a845aa-Paper.pdf}
  {Attention is all you need}.
\newblock In \emph{Advances in Neural Information Processing Systems},
  volume~30. Curran Associates, Inc.

\bibitem[{Wenzek et~al.(2020)Wenzek, Lachaux, Conneau, Chaudhary, Guzm{\'a}n,
  Joulin, and Grave}]{wenzek-etal-2020-ccnet}
Guillaume Wenzek, Marie-Anne Lachaux, Alexis Conneau, Vishrav Chaudhary,
  Francisco Guzm{\'a}n, Armand Joulin, and Edouard Grave. 2020.
\newblock \href {https://aclanthology.org/2020.lrec-1.494} {{CCN}et: Extracting
  high quality monolingual datasets from web crawl data}.
\newblock In \emph{Proceedings of the Twelfth Language Resources and Evaluation
  Conference}, pages 4003--4012, Marseille, France. European Language Resources
  Association.

\bibitem[{Wolf et~al.(2020)Wolf, Debut, Sanh, Chaumond, Delangue, Moi, Cistac,
  Rault, Louf, Funtowicz, Davison, Shleifer, von Platen, Ma, Jernite, Plu, Xu,
  Le~Scao, Gugger, Drame, Lhoest, and Rush}]{wolf-etal-2020-transformers}
Thomas Wolf, Lysandre Debut, Victor Sanh, Julien Chaumond, Clement Delangue,
  Anthony Moi, Pierric Cistac, Tim Rault, Remi Louf, Morgan Funtowicz, Joe
  Davison, Sam Shleifer, Patrick von Platen, Clara Ma, Yacine Jernite, Julien
  Plu, Canwen Xu, Teven Le~Scao, Sylvain Gugger, Mariama Drame, Quentin Lhoest,
  and Alexander Rush. 2020.
\newblock \href {https://doi.org/10.18653/v1/2020.emnlp-demos.6} {Transformers:
  State-of-the-art natural language processing}.
\newblock In \emph{Proceedings of the 2020 Conference on Empirical Methods in
  Natural Language Processing: System Demonstrations}, pages 38--45, Online.
  Association for Computational Linguistics.

\bibitem[{Yang et~al.(2019)Yang, Zhang, Tar, and
  Baldridge}]{yang-etal-2019-paws}
Yinfei Yang, Yuan Zhang, Chris Tar, and Jason Baldridge. 2019.
\newblock \href {https://doi.org/10.18653/v1/D19-1382} {{PAWS}-{X}: A
  cross-lingual adversarial dataset for paraphrase identification}.
\newblock In \emph{Proceedings of the 2019 Conference on Empirical Methods in
  Natural Language Processing and the 9th International Joint Conference on
  Natural Language Processing (EMNLP-IJCNLP)}, pages 3687--3692, Hong Kong,
  China. Association for Computational Linguistics.

\bibitem[{You et~al.(2020)You, Li, Reddi, Hseu, Kumar, Bhojanapalli, Song,
  Demmel, Keutzer, and Hsieh}]{You2020Large}
Yang You, Jing Li, Sashank Reddi, Jonathan Hseu, Sanjiv Kumar, Srinadh
  Bhojanapalli, Xiaodan Song, James Demmel, Kurt Keutzer, and Cho-Jui Hsieh.
  2020.
\newblock \href {https://openreview.net/forum?id=Syx4wnEtvH} {Large batch
  optimization for deep learning: Training bert in 76 minutes}.
\newblock In \emph{International Conference on Learning Representations}.

\end{thebibliography}
